\documentclass{article}

\usepackage[final,nonatbib]{nips_2017}

\usepackage{todonotes}
\usepackage[utf8]{inputenc} 
\usepackage[T1]{fontenc}    
\usepackage{hyperref}       
\usepackage{url}            
\usepackage{booktabs}       
\usepackage{amsfonts}       
\usepackage{nicefrac}       
\usepackage{microtype}      
\usepackage{tabularx, booktabs,adjustbox}
\usepackage{multirow}
\usepackage{caption}
\usepackage{subcaption}
\usepackage{array}
\usepackage{authblk}
\usepackage{float}
\usepackage{multirow}
\usepackage{footnote}
\usepackage[numbers]{natbib}

\begin{document}

\title{Development and Validation of Deep Learning Algorithms for Detection of Critical Findings in Head CT Scans}

\author[1]{Sasank Chilamkurthy}
\author[1]{Rohit Ghosh}
\author[1]{Swetha Tanamala}
\author[2]{Mustafa Biviji}
\author[3]{Norbert G. Campeau}
\author[4]{Vasantha Kumar Venugopal}
\author[4]{Vidur Mahajan}
\author[1]{Pooja Rao}
\author[1]{Prashant Warier}

\affil[1]{Qure.ai, Mumbai, IN}
\affil[2]{CT \& MRI Center, Nagpur, IN}
\affil[3]{Department of Radiology, Mayo Clinic, Rochester, MN}
\affil[4]{Centre for Advanced Research in Imaging, Neurosciences and Genomics, New Delhi, IN}

\newcommand{\absdiv}[1]{%
  \par\addvspace{0.5\baselineskip}
  \noindent\textbf{#1}\quad\ignorespaces
}

\maketitle
\begin{abstract}
\small
\absdiv{Importance}
Non-contrast head CT scan is the current standard for initial imaging of patients with head trauma or stroke symptoms.

\absdiv{Objective}
To develop and validate a set of deep learning algorithms for automated detection of following key findings from non-contrast head CT scans: intracranial hemorrhage (ICH) and its types, intraparenchymal (IPH), intraventricular (IVH), subdural (SDH), extradural (EDH) and subarachnoid (SAH) hemorrhages, calvarial fractures, midline shift and mass effect.

\absdiv{Design and Settings}
We retrospectively collected a dataset containing 313,318 head CT scans along with their clinical reports from various centers.
A part of this dataset (Qure25k dataset) was used to validate and the rest to develop algorithms.
Additionally, a dataset (CQ500 dataset) was collected from different centers in two batches B1 \& B2 to clinically validate the algorithms.

\absdiv{Main Outcomes and Measures}
Original clinical radiology report and consensus of three independent radiologists were considered as gold standard for Qure25k and CQ500 datasets respectively.
Area under receiver operating characteristics curve (AUC) for each finding was primarily used to evaluate the algorithms.

\absdiv{Results}
Qure25k dataset contained 21,095 scans (mean age $43.31$; $42.87\%$ female) while batches B1 and B2 of CQ500 dataset consisted of $214$ (mean age $43.40$; $43.92\%$ female) and $277$ (mean age $51.70$; $30.31\%$ female) scans respectively.
On Qure25k dataset, the algorithms achieved AUCs of 
$0.9194$,
$0.8977$,
$0.9559$,
$0.9161$,
$0.9288$ and
$0.9044$
for detecting ICH, IPH, IVH, SDH, EDH and SAH respectively.
AUCs for the same on CQ500 dataset were
$0.9419$,
$0.9544$,
$0.9310$,
$0.9521$,
$0.9731$ and
$0.9574$
respectively.
For detecting calvarial fractures, midline shift and mass effect, AUCs on Qure25k dataset were 
$0.9244$,
$0.9276$ and
$0.8583$
respectively, while AUCs on CQ500 dataset were
$0.9624$,
$0.9697$
and $0.9216$
respectively.

\absdiv{Conclusions and Relevance}
This study demonstrates that deep learning algorithms can accurately identify head CT scan abnormalities requiring urgent attention.
This opens up the possibility to use these algorithms to automate the triage process.
They may also provide a lower bound for quality and consistency of radiological interpretation.

\end{abstract}

\section{Introduction}

Non-contrast head CT scans are among the most commonly used emergency room diagnostic tools for patients with head injury or in those with symptoms suggesting a stroke or rise in intracranial pressure. Their wide availability and relatively low acquisition time makes them a commonly used first-line diagnostic modality \cite{ringl2010skull,doi:10.1093/bja/aem141}.
The percentage of annual US emergency room visits that involve a CT scan has been increasing for the last few decades \cite{larson2011national} and the use of head CT to exclude the need for neurosurgical intervention is on the rise\cite{papa2012performance}.

The most critical, time-sensitive abnormalities that can be readily detected on CT scan include intracranial hemorrhages, raised intracranial pressure and cranial fractures.
A key evaluation goal in patients with stroke is excluding an intracranial hemorrhage.
This depends on CT imaging and its swift interpretation.
Similarly, immediate CT scan interpretation is crucial in patients with a suspected acute intracranial hemorrhage to evaluate the need for neurosurgical treatment.
Cranial fractures, if open or depressed will usually require urgent neurosurgical intervention.
Cranial fractures are also the most commonly missed major abnormality on head CT scans \cite{wysoki1998head, erly2002radiology}, especially if coursing in an axial plane.

While these abnormalities are found only on a small fraction of CT scans, streamlining the head CT scan interpretation workflow by automating the initial screening and triage process, would significantly decrease the time to diagnosis and expedite treatment.
This would in turn decrease morbidity and mortality consequent to stroke and head injury.
An automated head CT scan screening and triage system would be valuable for queue management in a busy trauma care setting, or to facilitate decision-making in remote locations without an immediate radiologist availability.

The past year has seen a number of advances in application of deep learning \cite{yuan2017automatic,wang2017chestx,rajpurkar2017chexnet,gao2017classification,grewal2017radnet,yao2017learning} for medical imaging interpretation tasks, with robust evidence that deep learning can perform specific medical imaging tasks including identifying and grading diabetic retinopathy \cite{gulshan2016development} and classifying skin lesions as benign or malignant \cite{esteva2017dermatologist} with accuracy equivalent to specialist physicians.
Deep learning algorithms have also been trained to detect abnormalities on radiological images such as chest radiographs\cite{wang2017chestx,rajpurkar2017chexnet}, chest CT \cite{anthimopoulos2016lung,cheng2016computer} and head CT \cite{gao2017classification,grewal2017radnet} through `classification' algorithms; as well as to localize and quantify disease patterns or anatomical volumes \cite{prasoon2013deep,liao2013representation,patravali20172d} through `segmentation' algorithms.

The development of an accurate deep learning algorithm for radiology requires – in addition to appropriate model architectures – a large number of accurately labeled scans that will be used to train the algorithm\cite{litjens2017survey}.
The chances that the algorithm generalizes well to new settings increase when the training dataset is large and includes scans from diverse sources\cite{sun2017revisiting,halevy2009unreasonable} .

In this manuscript, we describe the development, validation and clinical testing of fully automated deep learning algorithms that are trained to detect abnormalities requiring urgent attention from non-contrast head CT scans.
The trained algorithms detect five kinds of intracranial hemorrhages (ICH) namely intraparenchymal (IPH), intraventricular (IVH), subdural (SDH), extradural (EDH) and subarachnoid (SAH), and calvarial/cranial vault fractures.
Algorithms also detect mass effect and midline shift, both used as indicators of severity of the brain injury.

\section{Methods}

\subsection{Datasets}
\label{sec:dataset}

We retrospectively collected 313,318 anonymous head CT scans from several centers in India.
These centers, which included both in-hospital and outpatient radiology centers, employ a variety of CT scanner models (shown in Table \ref{table:models}) with slices per rotation ranging from 2 to 128.
Each of the scans had a clinical report associated with them which we used as the gold standard during the algorithm development process.

Of these scans, we earmarked scans of 23,163 randomly selected patients (Qure25k dataset) for validation and used the scans of rest of the patients (development dataset) to train/develop the algorithms.
We removed post-operative scans and scans of patients less than 7 years old from the Qure25k dataset.
This dataset was \emph{not} used during the algorithm development process.

A clinical validation dataset (referred as CQ500 dataset) was provided by the Centre for Advanced Research in Imaging, Neurosciences and Genomics (CARING), New Delhi, India.
This dataset was a subset of head CT scans taken at various radiology centers in New Delhi.
Approximately, half of the centers are stand-alone outpatient centers and the other half are radiology departments embedded in large hospitals.
There was no overlap between these centers and the centers from where the development dataset was obtained.
CT scanners used at these centers had slices per rotation varying from 16 to 128.
Models of the CT scanners are listed in Table \ref{table:models}.
The data was pulled from local PACS servers and anonymized in compliance with internally defined HIPAA\cite{centers2003hipaa} guidelines.

\begin{table}
	\centering
	\begin{tabularx}{0.8\textwidth}{ l  X }
	\toprule
	\textbf{Dataset} & \textbf{CT Scanner Models} \\
	\midrule
	Qure25k \& Development & GE BrightSpeed, GE Brivo CT315, GE Brivo CT385, GE HiSpeed, GE LightSpeed, GE ProSpeed, GE Revolution ACTs, Philips Brilliance, Siemens Definition, Siemens Emotion, Siemens Sensation, Siemens SOMATOM, Siemens Spirit \\
	\midrule
	CQ500 & GE BrightSpeed, GE Discovery CT750 HD, GE LightSpeed, GE Optima CT660, Philips MX 16-slice, Philips Access-32 CT \\
	\bottomrule
	\end{tabularx}
	\bigskip
	\caption{Models of CT scanners used for each dataset}
	\label{table:models}
\end{table}

Similar to the development and Qure25k datasets, clinical radiology reports associated with the scans in the CQ500 dataset were available. Although, we do not use them as gold standards in this study, we use them for the dataset selection as described below.

We collected the CQ500 dataset in two batches (B1 \& B2).
Batch B1 was collected by selecting all the head CT scans taken at the above centers for 30 days starting from 20 November 2017. Batch B2 was selected from the rest of the scans in the following way:

\begin{enumerate}
	\item A natural language processing (NLP) algorithm was used to  detect IPH, SDH, EDH, SAH, IVH, calvarial fractures from clinical radiology reports.
	\item Reports were then randomly selected so that there are around 80 scans with each of IPH, SDH, EDH, SAH and calvarial fractures.
\end{enumerate}

Each of the selected scans were then screened for the following exclusion criteria:

\begin{itemize}
	\item Patient should not have any post-operative defect such as burr hole/shunt/clips etc.
	\item There should be at least one non-contrast CT series with axial cuts and soft reconstruction kernel which covers the complete brain.
	\item Patient should not be less than 7 years old. Wherever age information is not available, it is roughly estimated from ossification degree of cranial sutures\cite{harth2009estimating}.
\end{itemize}

Follow up scans for a patient were \emph{not} excluded in the selection process.

\subsection{Reading the scans} 

Three senior radiologists\footnote{Dr. Campeau was one of the three readers.} served as independent readers for the CT scans in the CQ500 dataset.
They had corresponding experience of 8, 12 and 20 years in cranial CT interpretation.
None of the 3 readers was involved in the clinical care or evaluation of the enrolled patients, nor did they have access to clinical history of any of the patients.
Each of the radiologists independently evaluated the scans in the CQ500 dataset with the instructions for recording the findings and query resolution as per the supplement.
The order of presentation of the scans was randomized so as to minimize the recall of the patients’ follow up scans.

CT scans were reviewed on a custom web based viewer built upon orthanc\cite{jodogne2013orthanc} and OHIF framework\cite{ohif}. Each reviewer could change the window level settings of the scans and had access to all the series present in a given CT scan.

Each of the readers recorded the following findings for each scan:
\begin{itemize}
	\item The presence or absence of a intracranial hemorrhage and if present, its type(s) (intraparenchymal, intraventricular, extradural, subdural and subarachnoid), age (chronic or not) and cerebral hemisphere(s) affected (left, right).
	\item The presence or absence of midline shift and mass effect.
	\item The presence or absence of fractures. If present, if it is (partly) a calvarial fracture.
\end{itemize}

\begin{figure}
	\centering
	\begin{subfigure}[t]{0.5\textwidth}
            \includegraphics[width=\linewidth]{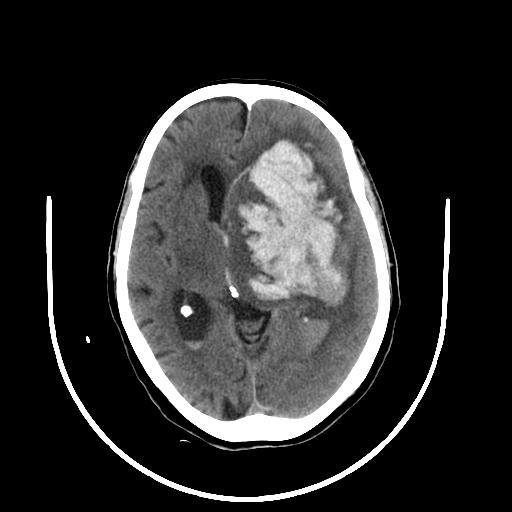}
            \caption{Slice from a head CT scan}
    \end{subfigure}
    \hspace{0.1\textwidth}
    \begin{subfigure}[t]{0.365\textwidth}
            \includegraphics[width=\linewidth]{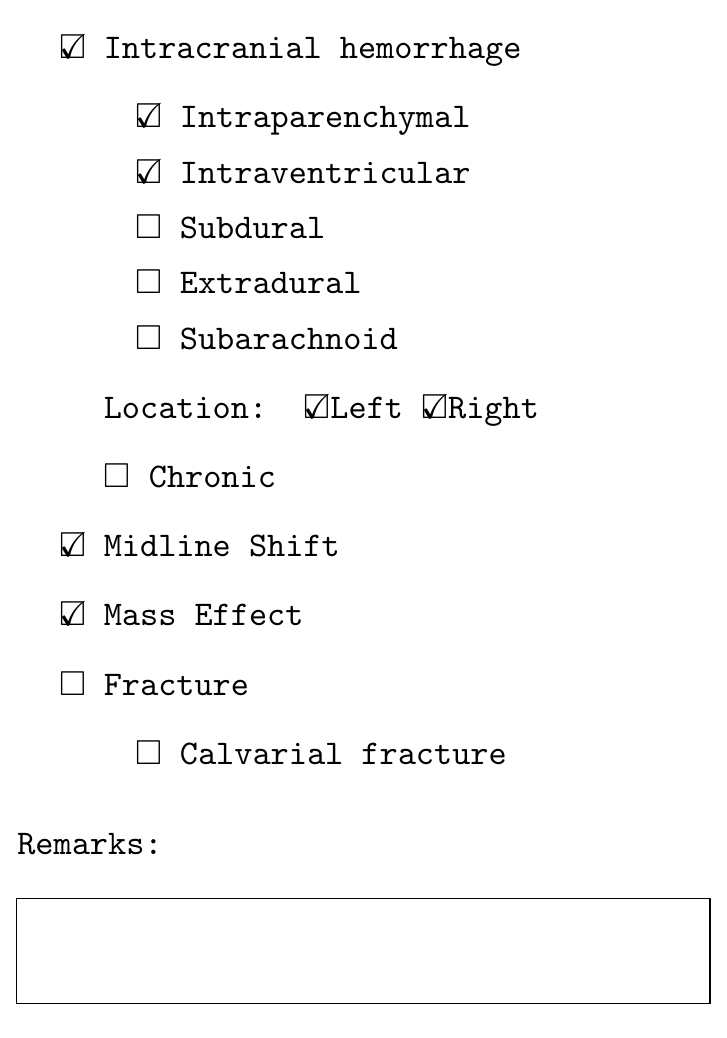}
		    \caption{Form used to record the findings}
    \end{subfigure}

    \caption{Schematic of the reading process for the CQ500 dataset.}
    \label{fig:reading}
\end{figure}

A text box was also provided to record any remark/observation which did not fit this decision flow. Schematic of the reading process along with the form used to record the findings is shown in Figure \ref{fig:reading}

Intra-axial presence of blood due to any etiology such as hemorrhagic contusion, tumour/infarct with hemorrhagic component was also included in the definition of intraparenchymal hemorrhage.
Chronic hemorrhages were considered positive in this study.
Mass effect was defined as any of the following: local mass effect, ventricular effacement, midline shift and herniation.
Midline shift was considered positive if the amount of shift was greater than 5mm.
If there is at least one fracture that extends into the skullcap, the scan is considered to have a calvarial fracture.

If unanimous agreement for each of the findings was not achieved by the three readers, the interpretation of majority of the readers was used as the final diagnosis.

On the development and Qure25k datasets, we considered clinical reports written by radiologists as the gold standard.
However, these were written in free text rather than in a structured format.
Therefore, a rule based natural language processing (NLP) algorithm was applied on radiologist’s clinical reports to automatically infer the findings recorded above.
We validated this algorithm on a subset of reports from the Qure25k dataset to ensure that the inferred information was accurate and could be used as gold standard.

\subsection{Developing the algorithms}
Deep learning is a form of machine learning where the model used is a neural network with a large number of (usually convolutional) layers\cite{deng2014deep}.
\emph{Training} this model requires large amount of data for which the ground truth is already known.
Training is usually performed by an algorithm called back propagation\cite{goodfellow2016deep,russell2003artificial}.
In this algorithm, model is iteratively modified to minimize the error between predictions of the model and the known ground truth for each data point.

One of the main challenges we faced in the development of the algorithms was the three dimensional (3D) nature of the CT scans. This was primarily due to an issue termed as `curse of dimensionality' where the data required to train a machine learning algorithm scales exponentially with the dimensionality of data \cite{domingos2012few}.
Deep learning techniques have been extensively researched for the tasks of
segmentation\cite{chen2016deeplab,long2015fully,ronneberger2015u} and classification\cite{krizhevsky2012imagenet,he2016deep} of two dimensional images. While the segmentation of 3D images is studied in multiple contexts \cite{cciccek20163d,kamnitsas2017efficient,patravali20172d}, their classification is not as well investigated.
One closely related problem is recognition of human actions from short video clips (because videos are three-dimensional with the time as the third dimension).
Despite that this problem is well explored in the literature \cite{carreira2017quo,simonyan2014two,ji20133d}, there was no emergence of a front running architecture for this task\cite{carreira2017quo}.
Our approach of classification is closely related to that of \citet{simonyan2014two} and involved slice level and pixel level annotation of a large number of scans.

In this study, we trained separate deep learning models for each of the subtasks viz. intracranial bleeds, midline shift/mass effect and calvarial fractures which we describe below.

\subsubsection{Intracranial Hemorrhage}

Development dataset was searched for non-contrast head CT scans which were reported with any of the IPH, IVH, SDH, EDH, SAH and those with neither of these.
Each slice in these scans was manually labeled with the hemorrhages that are visible in that slice.
In all, 4304 scans (165809 slices) were annotated, of which number of scans (slices) with IPH, IVH, SDH, EDH, SAH and neither of these were 1787 (12857), 299 (3147), 938 (11709), 623 (5424), 888 (6861) and 944 (133897) respectively.

We used ResNet18\cite{he2016deep}, a popular convolutional neural network architecture with a slight modification to predict softmax based confidences\cite{bridle1990probabilistic} for the presence of each type of hemorrhage in a slice.
We modified the architecture by using five parallel fully connected (FC) layers in place of a single FC layer.
This design was based on the assumption that the image features\cite{yosinski2014transferable} for detecting hemorrhages would be similar for all the hemorrhage types.
The confidences at the slice-level are combined using a random forest\cite{breiman2001random} to predict the scan-level confidence for the presence of intracranial hemorrhage and its types.

We further trained a model to localize the following type of hemorrhages: IPH, SDH, EDH.
Localization requires dense prediction\cite{long2015fully} of presence or absence of bleed for every pixel in the scan.
To train models for dense predictions, pixels corresponding to the each bleed were annotated for a subset of the above slice-annotated images to provide the ground truth for the model.
This set contained 1706 images of which number of images with IPH, SDH, EDH and neither of these are 506, 243, 277 and 750 respectively.
We used a UNet\cite{ronneberger2015u} based architecture for segmentation of each type of hemorrhage.

\subsubsection{Midline Shift and Mass Effect}

The algorithm for detecting midline shift and mass effect was very similar to the one for detecting intracranial hemorrhage.
Each slice from select scans was labeled for the presence or absence of midline shift and mass effect in that slice.
Overall, 699 scans (26135 slices) were annotated, of which number of scans (slices) with mass effect were 320 (3143) and midline shift were 249 (2074).

We used modified ResNet18 with two parallel fully connected layers to predict slice wise confidences for the presence of mass effect and midline shift respectively.
These slice level confidences were thereby combined using a random forest to predict scan-level confidences for both the abnormalities.

\subsubsection{Calvarial Fractures}

Development dataset was searched for scans with calvarial fractures. Each slice in these scans was annotated by marking a tight bounding box around fractures. Number of scans annotated were 1119 (42942 slices) of which 9938 slices showed a calvarial fracture.

Slices along with target bounding box was fed into a DeepLab\cite{chen2016deeplab} based architecture to predict pixel-wise heatmap for fractures.
Skull fractures are extremely sparse in this representation.
Gradient flow in the backpropagation algorithm tends to be hindered for such sparse signals.
We therefore employed hard negative mining loss \cite{felzenszwalb2010object,dalal2005histograms} to counter the sparsity of the annotation.

We engineered features representative of local fracture lesions and their volumes from the generated heatmaps of the whole scan.
We then used these features to train a random forest to predict scan-wise confidence of presence of a calvarial fracture.

\subsubsection{Preprocessing}

For a given CT scan, we selected the non-contrast axial series which uses soft reconstruction kernel and resampled it so that slice thickness is around 5mm.
We then resized all the slices of this series to a size of \(224 \times 224\) pixels before passing to our deep learning models.
Instead of passing the whole dynamic range of CT densities as a single channel, we windowed the densities by using three separate windows and stacking them as channels.
Windows used were brain window (\(l=40,w=80\)), bone window (\(l=500,w=3000\)) and subdural window (\(l=175,w=50\)).
This was because fracture visible in the bone window could indicate existence of an extra axial bleed in the brain window and conversely, presence of scalp hematoma in the the brain window could correlate with a fracture.
Subdural window helps differentiate between the skull and an extra axial bleed that might have been indistinguishable in a normal brain window\cite{brant2012fundamentals}.

\subsection{Evaluating the algorithms}

The combined algorithms when run on a scan produces a list of 9 real valued confidence scores in the range of [0, 1] indicating the presence of the following findings: Intracranial hemorrhage and each of the 5 types of hemorrhages, midline shift, mass effect and calvarial fracture.
As described before, the corresponding gold standards were obtained using majority voting for CQ500 dataset and by NLP algorithm of reports for Qure25k dataset.

For both CQ500 and Qure25k datasets, receiver operating characteristic (ROC) curves\cite{hanley1982meaning} were obtained for each of the above by varying the threshold and plotting true positive rate (i.e sensitivity) and false positive rate (i.e 1 - specificity) at each threshold.
Two operating points were chosen on the ROC curve so that sensitivity $\approx 0.9$ (high sensitivity point) and specificity $\approx 0.9$ (high specificity point) respectively. 
Areas under the ROC curve (AUCs) and sensitivities \& specificities at these two operating points were used to evaluate the algorithms.

\subsection{Statistical Analysis}
Sample sizes for proportions and AUCs were calculated using normal approximation and the method outlined by \citet{hanley1982meaning} respectively. 
The prevalence of our target abnormalities in a randomly selected sample of CT scans tend to be low.
This means that establishing the algorithms' sensitivity with a reasonably high confidence on an un-enriched dataset would require very large sample sizes.
For example, to establish a sensitivity with an expected value of $0.7$ within a 95\% confidence interval of half length of $0.10$, number of positive scans to be read $\approx 80$.
Similarly, for a finding with prevalence rate of $1\%$, to establish AUC within a 95\% confidence interval of half length of $0.05$, number of scans to be read $\approx 20000$.

The Qure25k dataset used in this study was randomly sampled from the population distribution and had number of scans $> 20000$ following the above sample size calculations.
However, constraints on the radiologist time necessitated the  selective sampling strategy outlined in the section \ref{sec:dataset} for the CQ500 dataset.
Manual curation of scans (by referring to the scans themselves) would have had selection bias towards more significant positive scans.
We mitigated this issue by random selection where positive scans were determined from the clinical reports.

We generated confusion matrices for each finding at the selected operating points.
We then calculated the $95\%$ confidence intervals for sensitivity and specificity from these matrices using ‘exact’ Clopper-Pearson method\cite{clopper1934use} based on Beta distribution.
Confidence intervals of area under the ROCs were calculated following the `distribution-based' approach described by \citet{hanley1982meaning}.
On the CQ500 dataset, we measured the concordance between paired readers on each finding using percentage of agreement and the Cohen's kappa(\(\kappa\)) statistic\cite{viera2005understanding}. In addition, we measured concordance between all the three readers on each finding using Fleiss' kappa(\(\kappa\)) \cite{fleiss1971measuring} statistic.

Sample sizes, sample statistics and confidence intervals were calculated using scikit-learn\cite{scikit-learn} and statsmodels\cite{seabold2010statsmodels} python packages.

\section{Results}

Patient demographics and prevalences for each finding are summarized in Table \ref{table:demographics}.
Qure25k dataset contained 21095 scans of which number of scans reported positive for intracranial hemorrhage and calvarial fracture are 2494 and 992 respectively.
CQ500 dataset included 491 scans of which batch B1 had 214 scans and batch B2 had 277 scans.
B1 contained 35 and 6 scans reported with intracranial hemorrhage and calvarial fracture respectively.
The same for B2 were 170 and 28 respectively.

\begin{table}
	\centering
	\begin{tabularx}{\textwidth}{p{17em} X X X}
	\toprule

	\textbf{Characteristic} & \textbf{Qure25k dataset} &
	\textbf{CQ500 dataset batch B1} & \textbf{CQ500 dataset batch B2} \\

	\midrule

	No. of scans & \(21095\) & \(214\) & \(277\) \\

	No. of readers per scan & \(1\) &  \(3\) & \(3\) \\

	\midrule
	\multicolumn{4}{l}{\textsc{Patient Demographics}} \\

	\multicolumn{4}{l}{Age} \\
	\hspace*{1em} No. of scans for which age was known & \(21095\) & \(189\) & \(251\)\\
	\hspace*{1em} Mean & \(43.31\) & \(43.40\) & \(51.70\) \\
	\hspace*{1em} Standard deviation & \(22.39\) & \(22.43\) & \(20.31\) \\
	\hspace*{1em} Range & \(7 - 99\) & \(7 - 95\) & \(10 - 95\) \\

	No. of females / No. of scans for which sex was known (percentage)  & 
	\(9030 / 21064\) \newline \small (\(42.87\)\%) &
	\(94 / 214\) \newline \small (\(43.92\)\%) &
	\(84 / 277\) \newline \small (\(30.31\)\%) \\

	\midrule
	\multicolumn{4}{l}{\textsc{Prevalence}} \\
	\multicolumn{4}{l}{No. of scans (percentage) with } \\
	\hspace*{1em} Intracranial hemorrhage
	& \(2494\) \small(\(11.82\)) &
    \(35\) \small (\(16.36\)) &
    \(170\) \small (\(61.37\)) \\

	\hspace*{3em} Intraparenchymal
	& \(2013\) \small(\(9.54\)) &
    \(29\) \small (\(13.55\)) &
    \(105\) \small (\(37.91\)) \\

	\hspace*{3em} Intraventricular
	& \(436\) \small(\(2.07\)) &
    \(7\) \small (\(3.27\)) &
    \(21\) \small (\(7.58\)) \\

	\hspace*{3em} Subdural
	& \(554\) \small(\(2.63\)) &
    \(9\) \small (\(4.21\)) &
    \(44\) \small (\(15.88\)) \\

	\hspace*{3em} Extradural
	& \(290\) \small(\(1.37\)) &
    \(2\) \small (\(0.93\)) &
    \(11\) \small (\(3.97\)) \\

	\hspace*{3em} Subarachnoid
	& \(611\) \small(\(2.90\)) &
    \(9\) \small (\(4.21\)) &
    \(51\) \small (\(18.41\)) \\

	\hspace*{1em} Fractures
	& \(1653\) \small(\(7.84\)) &
    \(8\) \small (\(3.74\)) &
    \(31\) \small (\(11.19\)) \\

	\hspace*{3em} Calvarial Fractures
	& \(992\) \small(\(4.70\)) &
    \(6\) \small (\(2.80\)) &
    \(28\) \small (\(10.11\)) \\

	\hspace*{1em} Midline Shift
	& \(666\) \small(\(3.16\)) &
    \(18\) \small (\(8.41\)) &
    \(47\) \small (\(16.97\)) \\

	\hspace*{1em} Mass effect
	& \(1517\) \small(\(7.19\)) &
    \(28\) \small (\(13.08\)) &
    \(99\) \small (\(35.74\)) \\

	\bottomrule
	\end{tabularx}
	\bigskip
	\caption{Dataset characteristics for CQ500 and Qure25k datasets.}
	\label{table:demographics}
\end{table}

\begin{figure}
	\centering
	\begin{subfigure}[t]{0.49\textwidth}
			\caption*{\textsc{CQ500 Batch B1}}
            \label{fig:gull}
            \includegraphics[width=0.9\linewidth]{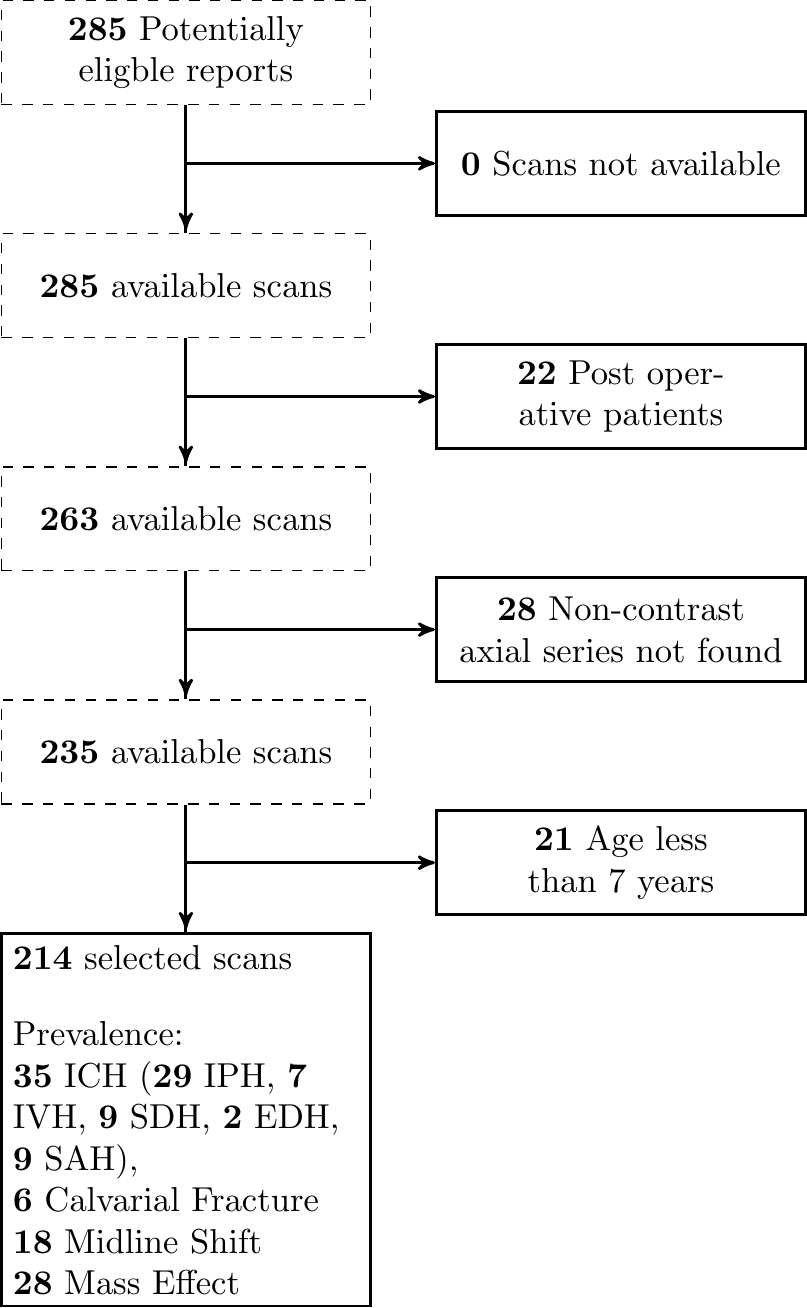}
    \end{subfigure}
    \begin{subfigure}[t]{0.49\textwidth}
		    \caption*{\textsc{CQ500 Batch B2}}
            \label{fig:gull2}
            \includegraphics[width=0.9\linewidth]{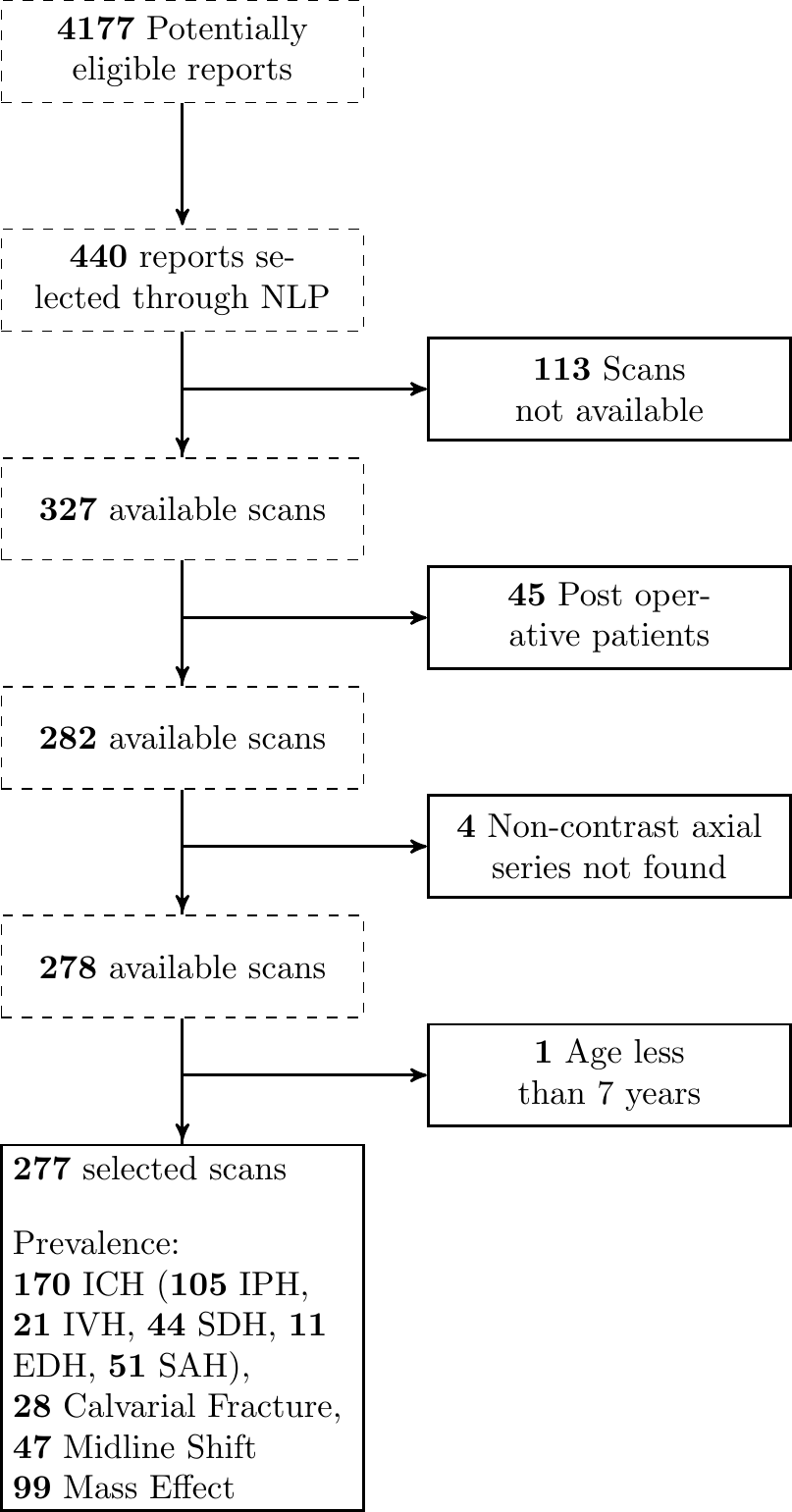}
    \end{subfigure}
    \bigskip
    \caption{Dataset section process for batches B1 and B2 of the CQ500 dataset.}
    \label{fig:studyselection}
\end{figure}

Number of clinical reports analyzed in the selection process of the CQ500 dataset was $4462$.
Of these, number of selected scans for batches B1 and B2 were $285$ and $440$ respectively. Number of exclusions were $71$ and $163$ respectively resulting in a total of $491$ scans.
Reasons for exclusion were non availability of images ($113$), post operative scans ($67$), scan had no non-contrast axial series ($32$) and patient was less than 7 years old ($22$).
Schematic of the dataset selection process of the CQ500 dataset is presented in Figure \ref{fig:studyselection}.

\subsection{Qure25k dataset}

Natural language processing (NLP) algorithm used to infer the findings from clinical reports in the Qure25k dataset was evaluated on a total of $1779$ reports.
Sensitivity and specificity of the NLP algorithm are fairly high; least performing finding was subdural hemorrhage with sensitivity of $0.9318$ ($95\%$ CI $0.8134$-$0.9857$) and specificity of $0.9965$ ($95\%$ CI $0.9925$-$0.9987$) while fracture was inferred perfectly with sensitivity of $1$ ($95\%$ CI $0.9745$-$1.000$) and specificity of $1$ ($95\%$ CI $0.9977$-$1.000$).
Sensitivity and specificity for all the target findings on the evaluated 1779 reports is shown in Table \ref{table:nlp}.

Table \ref{table:AUCqure25k} and Figure \ref{fig:ROC} summarize the performance of deep learning algorithms on the Qure25k set.
Algorithm achieved AUCs of $0.9194$ ($95\%$ CI $0.9119$-$0.9269$) on intracranial hemorrhage, $0.9244$ ($95\%$ CI $0.9130$-$0.9359$) on calvarial fracture and $0.9276$ ($95\%$ CI $0.9139$-$0.9413$) on midline shift. 
Algorithms performed the best on intraventricular hemorrhage with an AUC of $0.9559$ ($95\%$ CI $0.9424$-$0.9694$) and the worst on mass effect with an AUC of $0.8583$ ($95\%$ CI $0.8462$-$0.8703$).

At the high sensitivity operating point, sensitivities of the algorithms
for intracranial hemorrhage, calvarial fracture and midline shift were
$0.9006$ ($95\%$ CI $0.8882$-$0.9121$),
$0.9002$ ($95\%$ CI $0.8798$-$0.9181$) and
$0.9114$ ($95\%$ CI $0.8872$-$0.9319$)
respectively.
At this operating point, specificities for these findings were
$0.7295$ ($95\%$ CI $0.7230$-$0.7358$),
$0.7749$ ($95\%$ CI $0.7691$-$0.7807$) and
$0.8373$ ($95\%$ CI $0.8322$-$0.8424$)
respectively.

\begin{table}
	\begin{subtable}{\textwidth}
		\centering

		\begin{tabularx}{\textwidth}{ p{10em} p{5em} XX }
		\toprule

		\textbf{Finding} & \textbf{\#Positives} &
		\textbf{Sensitivity \newline \small($95\%$ CI)} &
		\textbf{Specificity \newline \small($95\%$ CI)} \\

		\midrule
		Intracranial Hemorrhage & \(207\) &
		\(0.9807\) \newline \tiny{(\(0.9513\)-\(0.9947\))} &
		 \(0.9873\) \newline \tiny{(\(0.9804\)-\(0.9922\))} \\

		\midrule
		\hspace*{2em}Intraparenchymal & \(157\) &
		\(0.9809\) \newline \tiny{(\(0.9452\)-\(0.9960\))} &
		\(0.9883\) \newline \tiny{(\(0.9818\)-\(0.9929\))} \\

		\midrule
		\hspace*{2em}Intraventricular & \(44\) &
		\(1.0000\) \newline \tiny{(\(0.9196\)-\(1.0000\))} &
		\(1.0000\) \newline \tiny{(\(0.9979\)-\(1.0000\))} \\

		\midrule
		\hspace*{2em}Subdural & \(44\) &
		\(0.9318\) \newline \tiny{(\(0.8134\)-\(0.9857\))} &
		\(0.9965\) \newline \tiny{(\(0.9925\)-\(0.9987\))} \\

		\midrule
		\hspace*{2em}Extradural & \(27\) &
		\(1.0000\) \newline \tiny{(\(0.8723\)-\(1.0000\))} &
		\(0.9983\) \newline \tiny{(\(0.9950\)-\(0.9996\))} \\

		\midrule
		\hspace*{2em}Subarachnoid & \(51\) &
		\(1.0000\) \newline \tiny{(\(0.9302\)-\(1.0000\))} &
		\(0.9971\) \newline \tiny{(\(0.9933\)-\(0.9991\))} \\

		\midrule
		Fracture & \(143\) &
		\(1.0000\) \newline \tiny{(\(0.9745\)-\(1.0000\))} &
		\(1.0000\) \newline \tiny{(\(0.9977\)-\(1.0000\))} \\

		\midrule
		\hspace*{2em}Calvarial Fracture & \(89\) &
		\(0.9888\) \newline \tiny{(\(0.9390\)-\(0.9997\))} &
		\(0.9947\) \newline \tiny{(\(0.9899\)-\(0.9976\))} \\

		\midrule
		Midline Shift & \(54\) &
		\(0.9815\) \newline \tiny{(\(0.9011\)-\(0.9995\))} &
		\(1.0000\) \newline \tiny{(\(0.9979\)-\(1.0000\))} \\

		\midrule
		Mass Effect & \(132\) &
		\(0.9773\) \newline \tiny{(\(0.9350\)-\(0.9953\))} &
		\(0.9933\) \newline \tiny{(\(0.9881\)-\(0.9967\))} \\

		\bottomrule
		\end{tabularx}
		\smallskip
		\caption{Qure25k dataset: performance of the NLP algorithm in inferring findings from the reports. This is measured on $1779$ reports from the Qure25k dataset}
		\label{table:nlp}
	\end{subtable}

	\bigskip
	\begin{subtable}{\textwidth}
		\centering
		\begin{tabularx}{\textwidth}{l XX XX XX X}
			\toprule

			\textbf{Finding} &
			\multicolumn{2}{c}{\textbf{Reader 1 \& 2}} &
			\multicolumn{2}{c}{\textbf{Reader 2 \& 3}} &
			\multicolumn{2}{c}{\textbf{Reader 3 \& 1}} &
			\textbf{All} \\

			&
			\small{Agreement \%} & \small{Cohen's \(\kappa\)} &
			\small{Agreement \%} & \small{Cohen's \(\kappa\)} &
			\small{Agreement \%} & \small{Cohen's \(\kappa\)} &
			\small{Fleiss' \(\kappa\)}
			\\
			\midrule

			Intracranial hemorrhage &
			\(89.00\) & \(0.7772\) &
			\(90.84\) & \(0.8084\) &
			\(88.39\) & \(0.7646\) &
			\(0.7827\) \\

			\midrule
			\hspace*{2em}Intraparenchymal &
			\(91.24\) & \(0.7865\) &
			\(90.63\) & \(0.7651\) &
			\(90.84\) & \(0.7719\) &
			\(0.7746\) \\

			\midrule
			\hspace*{2em}Intraventricular &
			\(96.13\) & \(0.7042\) &
			\(97.15\) & \(0.7350\) &
			\(95.72\) & \(0.6550\) &
			\(0.6962\) \\

			\midrule
			\hspace*{2em}Subdural &
			\(87.98\) & \(0.4853\) &
			\(93.08\) & \(0.6001\) &
			\(90.02\) & \(0.5624\) &
			\(0.5418\) \\

			\midrule
			\hspace*{2em}Extradural &
			\(97.35\) & \(0.5058\) &
			\(98.37\) & \(0.7251\) &
			\(98.17\) & \(0.5995\) &
			\(0.6145\) \\

			\midrule
			\hspace*{2em}Subarachnoid &
			\(93.08\) & \(0.6778\) &
			\(90.84\) & \(0.6058\) &
			\(90.84\) & \(0.6363\) &
			\(0.6382\) \\

			\midrule
			Calvarial Fracture &
			\(91.85\) & \(0.5771\) &
			\(92.06\) & \(0.3704\) &
			\(91.24\) & \(0.3637\) &
			\(0.4507\) \\

			\midrule
			Midline shift &
			\(88.19\) & \(0.5804\) &
			\(87.17\) & \(0.5344\) &
			\(93.69\) & \(0.7036\) &
			\(0.5954\) \\

			\midrule
			Mass effect &
			\(86.35\) & \(0.6541\) &
			\(87.98\) & \(0.6747\) &
			\(86.97\) & \(0.6837\) &
			\(0.6698\) \\

			\bottomrule
		\end{tabularx}
		\smallskip
		\caption{CQ500 dataset: concordance between the readers}
		\label{table:concordance}
	\end{subtable}

	\bigskip
	\caption{Reliability of the gold standards for Qure25k and CQ500 datasets.
			 On Qure25k, we used an NLP algorithm to infer findings from a radiologist’s report.
			 Three radiologists reviewed each of the 491 cases on CQ500 dataset and majority vote of the readers is used as gold standard.
			 Table \ref{table:nlp} shows the estimates of reliability of the used NLP algorithm while
			 Table \ref{table:concordance} shows the reliability and concordance of radiologists' reads}
\end{table}
\begin{table}
	\begin{subtable}{\textwidth}
		\centering
		\begin{tabularx}{\linewidth}{p{8.5em} p{4.9em} p{4.9em} p{4.9em} p{4.9em}p{4.9em}}
		\toprule

		\textbf{Finding} & \textbf{AUC \newline \small ($95\%$ CI)} &
		\multicolumn{2}{p{10em}}{\textbf{High sensitivity operating point}} &
		\multicolumn{2}{p{10em}}{\textbf{High specificity operating point}} \\

		 & &
		Sensitivity \newline \small ($95\%$ CI) &
		Specificity \newline \small ($95\%$ CI) &
		Sensitivity \newline \small ($95\%$ CI) &
		Specificity \newline \small ($95\%$ CI) \\

		\midrule
		Intracranial \newline hemorrhage &
		\(0.9194\) \newline \tiny{(\(0.9119\)-\(0.9269\))}  &
		\(0.9006\) \newline \tiny{(\(0.8882\)-\(0.9121\))} &
		\(0.7295\) \newline \tiny{(\(0.7230\)-\(0.7358\))} &
		\(0.8349\) \newline \tiny{(\(0.8197\)-\(0.8492\))} &
		\(0.9004\) \newline \tiny{(\(0.8960\)-\(0.9047\))} \\

		\midrule
		\hspace*{1em}Intraparenchymal &
		\(0.8977\) \newline \tiny{(\(0.8884\)-\(0.9069\))} &
		\(0.9031\) \newline \tiny{(\(0.8894\)-\(0.9157\))} &
		\(0.6046\) \newline \tiny{(\(0.5976\)-\(0.6115\))} &
		\(0.7670\) \newline \tiny{(\(0.7479\)-\(0.7853\))} &
		\(0.9046\) \newline \tiny{(\(0.9003\)-\(0.9087\))} \\

		\midrule
		\hspace*{1em}Intraventricular &
		\(0.9559\) \newline \tiny{(\(0.9424\)-\(0.9694\))} &
		\(0.9358\) \newline \tiny{(\(0.9085\)-\(0.9569\))} &
		\(0.8343\) \newline \tiny{(\(0.8291\)-\(0.8393\))} &
		\(0.9220\) \newline \tiny{(\(0.8927\)-\(0.9454\))} &
		\(0.9267\) \newline \tiny{(\(0.9231\)-\(0.9302\))} \\

		\midrule
		\hspace*{1em}Subdural &
		\(0.9161\) \newline \tiny{(\(0.9001\)-\(0.9321\))} &
		\(0.9152\) \newline \tiny{(\(0.8888\)-\(0.9370\))} &
		\(0.6542\) \newline \tiny{(\(0.6476\)-\(0.6607\))} &
		\(0.7960\) \newline \tiny{(\(0.7600\)-\(0.8288\))} &
		\(0.9041\) \newline \tiny{(\(0.9000\)-\(0.9081\))} \\

		\midrule
		\hspace*{1em}Extradural &
		\(0.9288\) \newline \tiny{(\(0.9083\)-\(0.9494\))} &
		\(0.9034\) \newline \tiny{(\(0.8635\)-\(0.9349\))} &
		\(0.7936\) \newline \tiny{(\(0.7880\)-\(0.7991\))} &
		\(0.8207\) \newline \tiny{(\(0.7716\)-\(0.8631\))} &
		\(0.9068\) \newline \tiny{(\(0.9027\)-\(0.9107\))} \\

		\midrule
		\hspace*{1em}Subarachnoid &
		\(0.9044\) \newline \tiny{(\(0.8882\)-\(0.9205\))} &
		\(0.9100\) \newline \tiny{(\(0.8844\)-\(0.9315\))} &
		\(0.6678\) \newline \tiny{(\(0.6613\)-\(0.6742\))} &
		\(0.7758\) \newline \tiny{(\(0.7406\)-\(0.8083\))} &
		\(0.9012\) \newline \tiny{(\(0.8971\)-\(0.9053\))} \\

		\midrule
		Calvarial fracture &
		\(0.9244\) \newline \tiny{(\(0.9130\)-\(0.9359\))} &
		\(0.9002\) \newline \tiny{(\(0.8798\)-\(0.9181\))} &
		\(0.7749\) \newline \tiny{(\(0.7691\)-\(0.7807\))} &
		\(0.8115\) \newline \tiny{(\(0.7857\)-\(0.8354\))} &
		\(0.9020\) \newline \tiny{(\(0.8978\)-\(0.9061\))} \\

		\midrule
		Midline Shift &
		\(0.9276\) \newline \tiny{(\(0.9139\)-\(0.9413\))} &
		\(0.9114\) \newline \tiny{(\(0.8872\)-\(0.9319\))} &
		\(0.8373\) \newline \tiny{(\(0.8322\)-\(0.8424\))} &
		\(0.8754\) \newline \tiny{(\(0.8479\)-\(0.8995\))} &
		\(0.9006\) \newline \tiny{(\(0.8964\)-\(0.9047\))} \\

		\midrule
		Mass Effect &
		\(0.8583\) \newline \tiny{(\(0.8462\)-\(0.8703\))} &
		\(0.8622\) \newline \tiny{(\(0.8439\)-\(0.8792\))} &
		\(0.6157\) \newline \tiny{(\(0.6089\)-\(0.6226\))} &
		\(0.7086\) \newline \tiny{(\(0.6851\)-\(0.7314\))} &
		\(0.9068\) \newline \tiny{(\(0.9026\)-\(0.9108\))} \\

		\bottomrule
		\end{tabularx}
		\smallskip
		\caption{Qure25k dataset: the algorithms' performance}
		\label{table:AUCqure25k}
	\end{subtable}

	\bigskip

	\begin{subtable}{\textwidth}
		\centering
		\begin{tabularx}{\textwidth}{ p{8.5em} p{4.9em} p{4.9em} p{4.9em} p{4.9em}p{4.9em}}
		\toprule

		\textbf{Finding} & \textbf{AUC \newline \small ($95\%$ CI)} &
		\multicolumn{2}{p{10em}}{\textbf{High sensitivity operating point}} &
		\multicolumn{2}{p{10em}}{\textbf{High specificity operating point}} \\

		 & &
		Sensitivity \newline \small ($95\%$ CI) &
		Specificity \newline \small ($95\%$ CI) &
		Sensitivity \newline \small ($95\%$ CI) &
		Specificity \newline \small ($95\%$ CI) \\

		\midrule
		Intracranial \newline hemorrhage &
		\(0.9419\) \newline \tiny{(\(0.9187\)-\(0.9651\))} &
		\(0.9463\) \newline \tiny{(\(0.9060\)-\(0.9729\))} &
		\(0.7098\) \newline \tiny{(\(0.6535\)-\(0.7617\))} &
		\(0.8195\) \newline \tiny{(\(0.7599\)-\(0.8696\))} &
		\(0.9021\) \newline \tiny{(\(0.8616\)-\(0.9340\))} \\

		\midrule
		\hspace*{1em}Intraparenchymal &
		\(0.9544\) \newline \tiny{(\(0.9293\)-\(0.9795\))} &
		\(0.9478\) \newline \tiny{(\(0.8953\)-\(0.9787\))} &
		\(0.8123\) \newline \tiny{(\(0.7679\)-\(0.8515\))} &
		\(0.8433\) \newline \tiny{(\(0.7705\)-\(0.9003\))} &
		\(0.9076\) \newline \tiny{(\(0.8726\)-\(0.9355\))} \\

		\midrule
		\hspace*{1em}Intraventricular &
		\(0.9310\) \newline \tiny{(\(0.8654\)-\(0.9965\))} &
		\(0.9286\) \newline \tiny{(\(0.7650\)-\(0.9912\))} &
		\(0.6652\) \newline \tiny{(\(0.6202\)-\(0.7081\))} &
		\(0.8929\) \newline \tiny{(\(0.7177\)-\(0.9773\))} &
		\(0.9028\) \newline \tiny{(\(0.8721\)-\(0.9282\))} \\

		\midrule
		\hspace*{1em}Subdural &
		\(0.9521\) \newline \tiny{(\(0.9117\)-\(0.9925\))} &
		\(0.9434\) \newline \tiny{(\(0.8434\)-\(0.9882\))} &
		\(0.7215\) \newline \tiny{(\(0.6769\)-\(0.7630\))} &
		\(0.8868\) \newline \tiny{(\(0.7697\)-\(0.9573\))} &
		\(0.9041\) \newline \tiny{(\(0.8726\)-\(0.9300\))} \\

		\midrule
		\hspace*{1em}Extradural &
		\(0.9731\) \newline \tiny{(\(0.9113\)-\(1.0000\))} &
		\(0.9231\) \newline \tiny{(\(0.6397\)-\(0.9981\))} &
		\(0.8828\) \newline \tiny{(\(0.8506\)-\(0.9103\))} &
		\(0.8462\) \newline \tiny{(\(0.5455\)-\(0.9808\))} &
		\(0.9477\) \newline \tiny{(\(0.9238\)-\(0.9659\))} \\

		\midrule
		\hspace*{1em}Subarachnoid &
		\(0.9574\) \newline \tiny{(\(0.9214\)-\(0.9934\))} &
		\(0.9167\) \newline \tiny{(\(0.8161\)-\(0.9724\))} &
		\(0.8654\) \newline \tiny{(\(0.8295\)-\(0.8962\))} &
		\(0.8667\) \newline \tiny{(\(0.7541\)-\(0.9406\))} &
		\(0.9049\) \newline \tiny{(\(0.8732\)-\(0.9309\))} \\

		\midrule
		Calvarial fracture &
		\(0.9624\) \newline \tiny{(\(0.9204\)-\(1.0000\))}&
		\(0.9487\) \newline \tiny{(\(0.8268\)-\(0.9937\))} &
		\(0.8606\) \newline \tiny{(\(0.8252\)-\(0.8912\))} &
		\(0.8718\) \newline \tiny{(\(0.7257\)-\(0.9570\))} &
		\(0.9027\) \newline \tiny{(\(0.8715\)-\(0.9284\))} \\

		\midrule
		Midline shift &
		\(0.9697\) \newline \tiny{(\(0.9403\)-\(0.9991\))} &
		\(0.9385\) \newline \tiny{(\(0.8499\)-\(0.9830\))} &
		\(0.8944\) \newline \tiny{(\(0.8612\)-\(0.9219\))} &
		\(0.9077\) \newline \tiny{(\(0.8098\)-\(0.9654\))} &
		\(0.9108\) \newline \tiny{(\(0.8796\)-\(0.9361\))} \\

		\midrule
		Mass effect &
		\(0.9216\) \newline \tiny{(\(0.8883\)-\(0.9548\))} &
		\(0.9055\) \newline \tiny{(\(0.8408\)-\(0.9502\))} &
		\(0.7335\) \newline \tiny{(\(0.6849\)-\(0.7782\))} &
		\(0.8189\) \newline \tiny{(\(0.7408\)-\(0.8816\))} &
		\(0.9038\) \newline \tiny{(\(0.8688\)-\(0.9321\))} \\

		\bottomrule
		\end{tabularx}
		\smallskip
		\caption{CQ500 dataset: the algorithms' performance}
		\label{table:AUC_CQ500}
	\end{subtable}
	\smallskip
	\caption{Performance of the algorithms on the Qure25k and CQ500 datasets. Neither of the datasets was used during the training process. AUCs are shown for 9 critical CT findings on both these datasets. Two operating points were chosen on the ROC for high sensitivity and high specificity respectively.}
\end{table}

\subsection{CQ500 dataset}

Concordance between the three readers on the CQ500 dataset was observed to be the highest for intracranial hemorrhage (Fleiss' \(\kappa=0.7827\)) and intraparenchymal hemorrhage  (Fleiss' \(\kappa=0.7746\)).
Calvarial fracture and subdural hemorrhage had the lowest concordance with Fleiss' \(\kappa\) of \(0.4507\) and \(0.5418\) respectively.
For each of the target findings, percentage agreement and Cohen's kappa between a pair of readers and Fleiss' kappa for all the readers is shown in \ref{table:concordance}

\begin{figure}
	\begin{subfigure}{0.32\textwidth}
	\includegraphics[width=\linewidth]{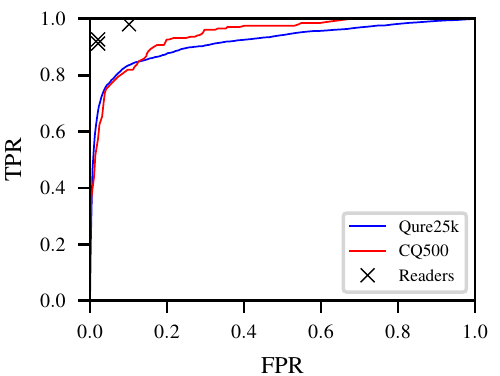}
	\caption*{\small Intracranial hemorrhage}
	\end{subfigure}
	\hfill
	\begin{subfigure}{0.32\textwidth}
	\includegraphics[width=\linewidth]{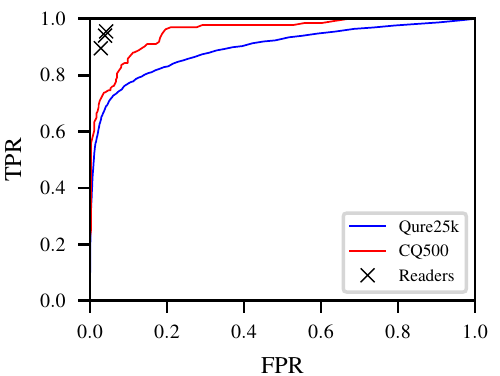}
	\caption*{\small Intraparenchymal hemorrhage}
	\end{subfigure}
	\hfill
	\begin{subfigure}{0.32\textwidth}
	\includegraphics[width=\linewidth]{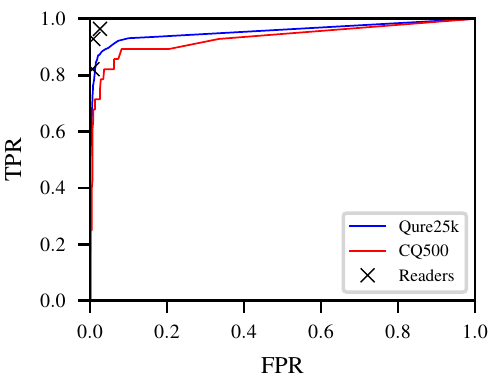}
	\caption*{\small Intraventricular hemorrhage}
	\end{subfigure}
	\bigskip

	\begin{subfigure}{0.32\textwidth}
	\includegraphics[width=\linewidth]{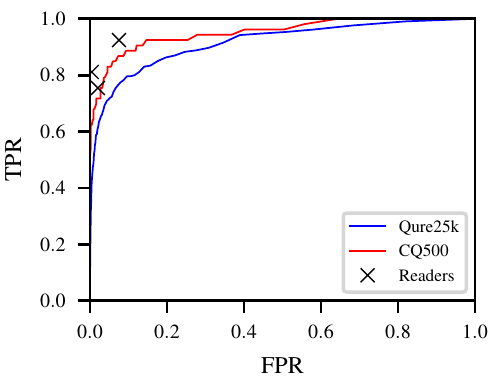}
	\caption*{\small Subdural hemorrhage}
	\end{subfigure}
	\hfill
	\begin{subfigure}{0.32\textwidth}
	\includegraphics[width=\linewidth]{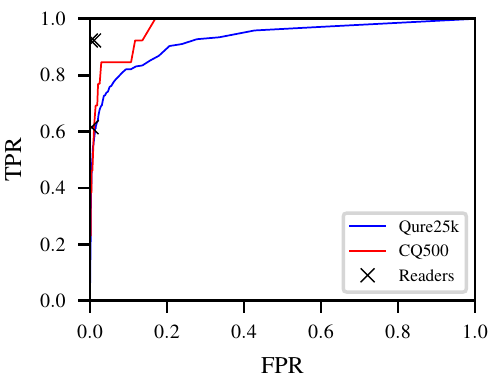}
	\caption*{\small Extradural hemorrhage}
	\end{subfigure}
	\hfill
	\begin{subfigure}{0.32\textwidth}
	\includegraphics[width=\linewidth]{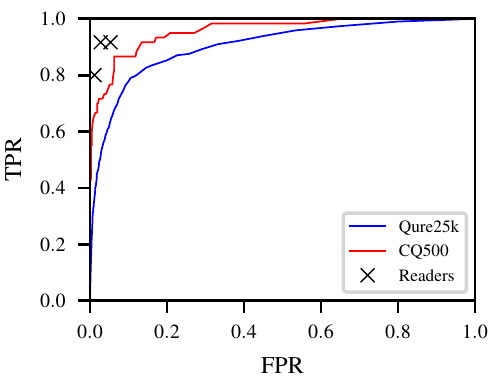}
	\caption*{\small Subarachnoid hemorrhage}
	\end{subfigure}
	\bigskip

	\begin{subfigure}{0.32\textwidth}
	\includegraphics[width=\linewidth]{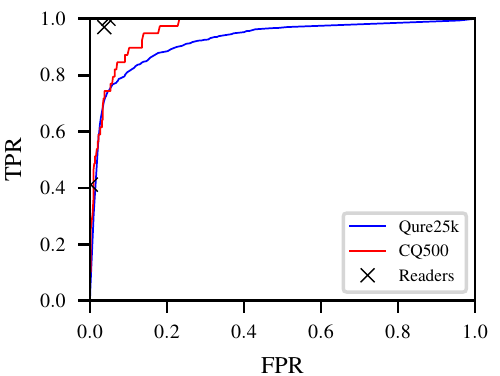}
	\caption*{\small Calvarial Fracture}
	\end{subfigure}
	\hfill
	\begin{subfigure}{0.32\textwidth}
	\includegraphics[width=\linewidth]{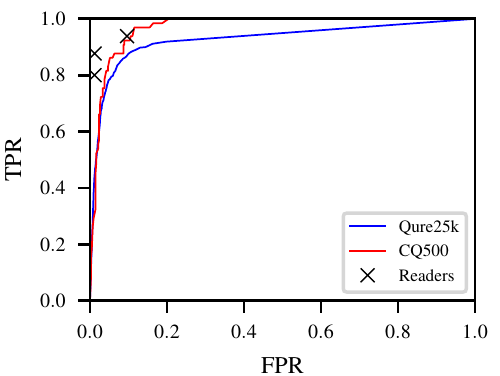}
	\caption*{\small Midline Shift}
	\end{subfigure}
	\hfill
	\begin{subfigure}{0.32\textwidth}
	\includegraphics[width=\linewidth]{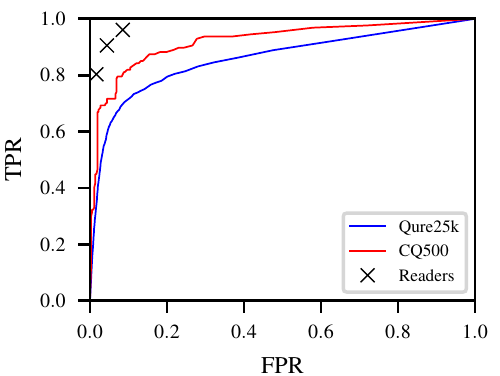}
	\caption*{\small Mass effect}
	\end{subfigure}
	\bigskip
	\caption{Receiver operating characteristic (ROC) curves for the algorithms on Qure25k and CQ500 datasets.
			 Blue lines are for the Qure25k dataset and Red lines are for the CQ500 dataset.
			 Readers' TPR and FPR against consensus on CQ500 dataset 
			 are plotted along with the ROCs for comparison}
	\label{fig:ROC}
\end{figure}

The algorithms generally performed better on the CQ500 dataset than on the Qure25k dataset.
AUCs, sensitivities and specificities are shown in Table \ref{table:AUC_CQ500} and ROCs are shown in \ref{fig:ROC}.
AUC for intracranial hemorrhage was
$0.9419$ ($95\%$ CI $0.9187$-$0.9651$), for calvarial fracture was
$0.9624$ ($95\%$ CI $0.9204$-$1.0000$), and for midline shift was
$0.9697$ ($95\%$ CI $0.9403$-$0.9991$).
Best AUCs were recorded on midline shift ($0.9697$ , $95\%$ CI $0.9403$-$0.9991$) and
extradural hemorrhage ($0.9731$, $95\%$ CI $0.9113$-$1.0000$)
while the worst AUC was on 
mass effect ($0.9216$, $95\%$ CI $0.8883$-$0.9548$).

Sensitivities of the algorithms at the high sensitivity operating point for intracranial hemorrhage, calvarial fracture and midline shift were
\(0.9463\) ($95\%$ CI \(0.9060\)-\(0.9729\)),
\(0.9487\) ($95\%$ CI \(0.8268\)-\(0.9937\)) and
\(0.9385\) ($95\%$ CI \(0.8499\)-\(0.9830\)) respectively.
Specificities for the same were
\(0.7098\) ($95\%$ CI \(0.6535\)-\(0.7617\)),
\(0.8606\) ($95\%$ CI \(0.8252\)-\(0.8912\)) and
\(0.8944\) ($95\%$ CI \(0.8612\)-\(0.9219\)) respectively.

Algorithms' performance difference between the CQ500 dataset and the Qure25k dataset was most amplified for mass effect (AUC of \(0.9216\) vs \(0.8583\)) and the least for intracranial hemorrhage (AUC of \(0.9419\) vs \(0.9149\)).

\section{Discussion}

Automated and semi automated detection of findings from head CT scans have been studied by other groups.
\citet{grewal2017radnet} developed a deep learning approach to automatically detect intracranial hemorrhages.
They reported a sensitivity of $0.8864$ and a positive predictive value (precision) of $0.8125$ on a dataset of 77 brain CT scans read by three radiologists.
However, the types of intracranial hemorrhage considered were not mentioned in their report.
Traditional computer vision techniques like morphological processing were used by \citet{zaki2009new} and \citet{yamada2016preliminary} to detect and retrieve scans with fracture respectively.
Neither of the two studies measured their accuracies on a clinical dataset.
Automated midline shift detection was also explored\cite{chen2013automated,wang2017simple,xiao2010automated} using non deep learning methods.
Convolutional neural networks were used by \citet{gao2017classification} to classify head CT scans to help diagnose Alzheimer's disease.
More recently, \citet{prevedello2017automated} evaluated the performance of a deep learning algorithm on a dataset of 50 scans to detect hemorrhage, mass effect, or hydrocephalus (HMH) and suspected acute infarct (SAI).
They reported AUCs of 0.91 and 0.81 on HMH and SAI respectively.

Our study is the first to describe the development of a system that separately identifies critical abnormalities on head CT scans and to conduct a validation with a large number (21095) of scans sampled uniformly from the population distribution.
We also report the algorithms' accuracy versus a consensus of 3 radiologists on a second independent dataset, the CQ500 dataset.
We make this dataset and the corresponding reads available for public access\footnote{CQ500 dataset is available for download at \url{http://headctstudy.qure.ai/dataset}}, so that they can be used to benchmark comparable algorithms in the future.
Such publicly available datasets had earlier spurred comparison of the algorithms in other tasks like lung nodule detection\cite{armato2011lung} and chest radiograph diagnosis\cite{wang2017chestx}.

This work is novel because it is the first large study describing the use of deep learning on head CT scans to detect and separately report accuracy on each finding, including the five types of intracranial hemorrhage.
Further, there is very little literature to date describing the accurate use of deep learning algorithms to detect cranial fractures - we demonstrate that deep learning algorithms are able to perform this task with high accuracy.
The clinical validation of algorithms that detect mass effect and midline shift (both used to estimate severity of a variety on intracranial conditions and the need for urgent intervention) on such a large number of patients is also unique.

The algorithms produced fairly good results for all the target findings on both the Qure25k and CQ500 datasets.
AUCs for all the findings except mass effect were greater than or approximately equal to $0.9$.
AUCs on CQ500 dataset were significantly better than that on the Qure25k dataset.
We hypothesize that this might be because of two reasons. 
Firstly, since radiologists reading Qure25k dataset had access to clinical history of the patient, their reads incorporated extra clinical information not available in the scans.
The algorithms didn't have access to this information and therefore did not perform well.
Secondly, majority vote of three readers is a better gold standard than that of one reader.
Indeed, we observed that AUCs of the algorithms on CQ500 were lower when a single reader was considered the gold standard instead of the majority vote (see Table \ref{table:single_reader_aucs}). 

\begin{table}
	\centering
	\begin{tabularx}{\textwidth}{p{10em} X X X X}
		\toprule

		\multicolumn{1}{r}{\textbf{Gold standard}}&
		Reader 1 &
		Reader 2 &
		Reader 3 &
		Majority Vote\\

		\textbf{Finding} & \\

		\midrule

		Intracranial hemorrhage &
		0.9080 &
		0.9413 &
		0.9356 &
		0.9419\\

		Calvarial fracture &
		0.9198 &
		0.8653 &
		0.9523 &
		0.9624\\

		Midline shift &
		0.9545 &
		0.9386 &
		0.9461 &
		0.9697 \\

		\bottomrule
	\end{tabularx}
	\bigskip
	\caption{AUCs of the algorithms on CQ500 dataset when a single reader is
			 considered the gold standard and when the majority vote is considered the gold standard}
	\label{table:single_reader_aucs}
	\vspace{-1em}
\end{table}

\begin{figure}
	\begin{subfigure}{\textwidth}
		\centering
		\begin{subfigure}{0.3\textwidth}
			\includegraphics[width=\textwidth]{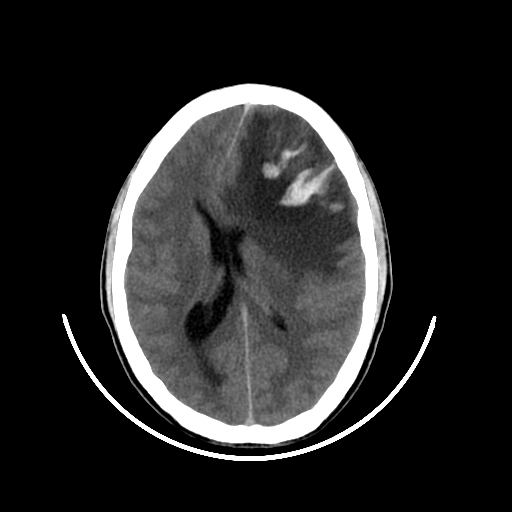}
			\caption{Intraparenchymal hemorrhage (in left frontal region)}
			\medskip
		\end{subfigure}
		\hfill
		\begin{subfigure}{0.3\textwidth}
			\includegraphics[width=\textwidth]{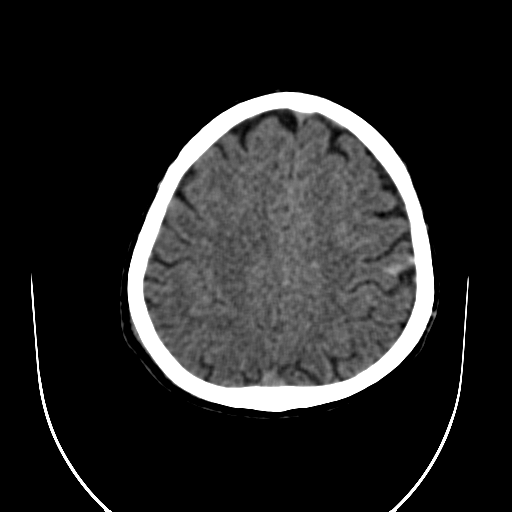}
			\caption{Subarachnoid hemorrhage (in left parietal region)}
			\medskip	
		\end{subfigure}
		\hfill
		\begin{subfigure}{0.3\textwidth}
			\includegraphics[width=\textwidth]{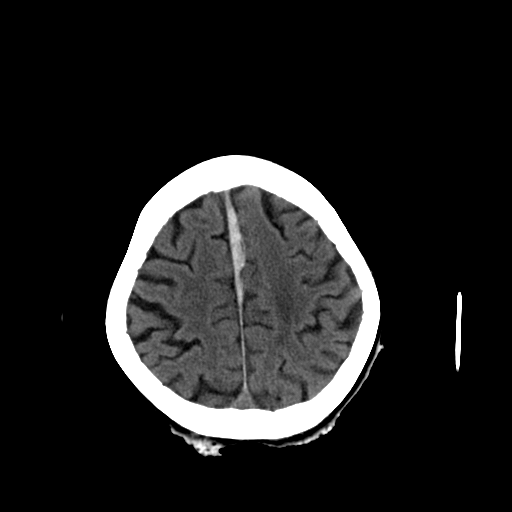}
			\caption{Subdural hemorrhage (along falx)}
			\medskip	
		\end{subfigure}
		\hfill
		\begin{subfigure}{0.3\textwidth}
			\includegraphics[width=\textwidth]{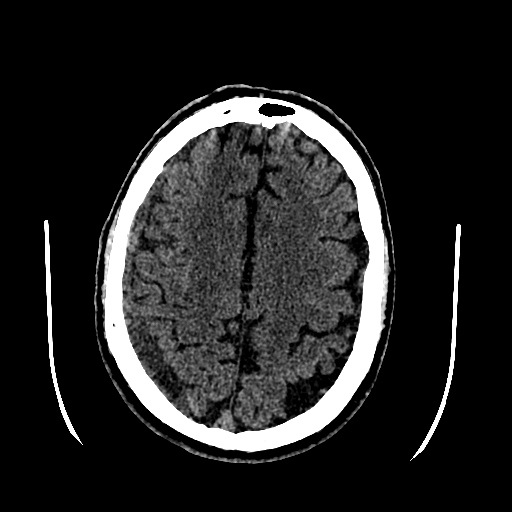}
			\caption{Subdural hemorrhage (chronic in right parietal convexity)}
			\medskip	
		\end{subfigure}
		\hfill
		\begin{subfigure}{0.3\textwidth}
			\includegraphics[width=\textwidth]{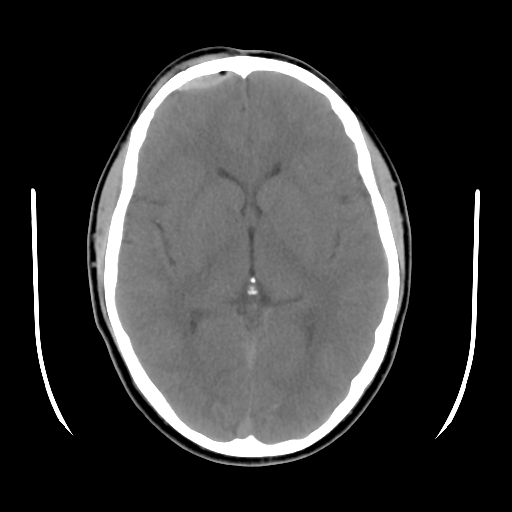}
			\caption{Extradural hemorrhage (in right frontal convexity)}
			\medskip	
		\end{subfigure}
		\hfill
		\begin{subfigure}{0.3\textwidth}
			\includegraphics[width=\textwidth]{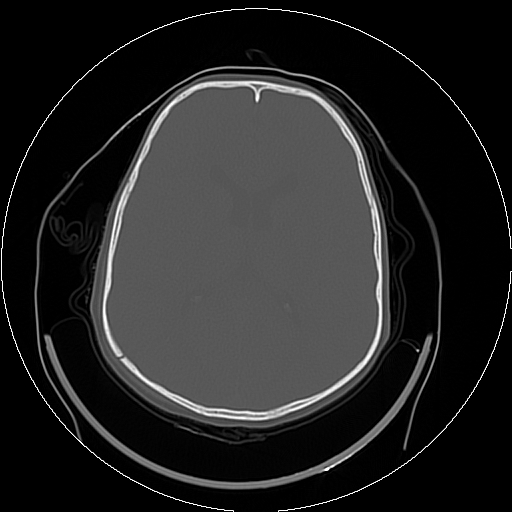}
			\caption{Calvarial fracture (in right parietal bone)}
			\medskip	
		\end{subfigure}
		\hfill
		\begin{subfigure}{0.3\textwidth}
			\includegraphics[width=\textwidth]{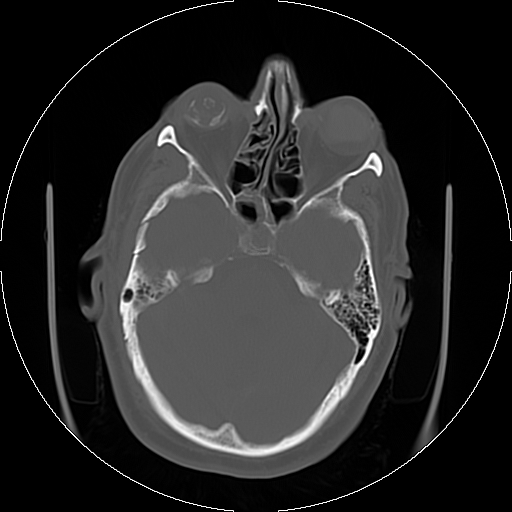}
			\caption{Calvarial fracture (in right temporal bone)}
		\end{subfigure}
		\hfill
		\begin{subfigure}{0.3\textwidth}
			\includegraphics[width=\textwidth]{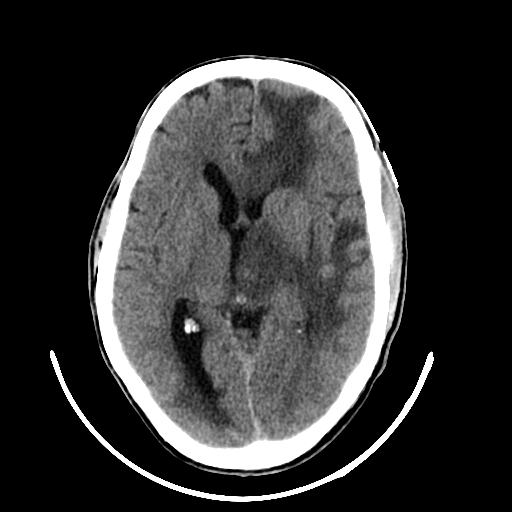}
			\caption{Midline shift\newline}
		\end{subfigure}
		\hfill
		\begin{subfigure}{0.3\textwidth}
			\includegraphics[width=\textwidth]{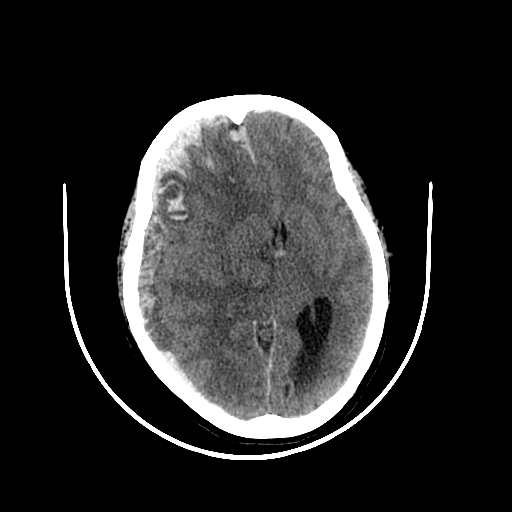}
			\caption{Midline shift\newline}
		\end{subfigure}
		\vspace{-3mm}
		\caption*{Accurate predictions: True Positives}
		\vspace{5mm}
	\end{subfigure}
	\begin{subfigure}{\textwidth}
		\centering
		\begin{subfigure}{0.3\textwidth}
			\includegraphics[width=\textwidth]{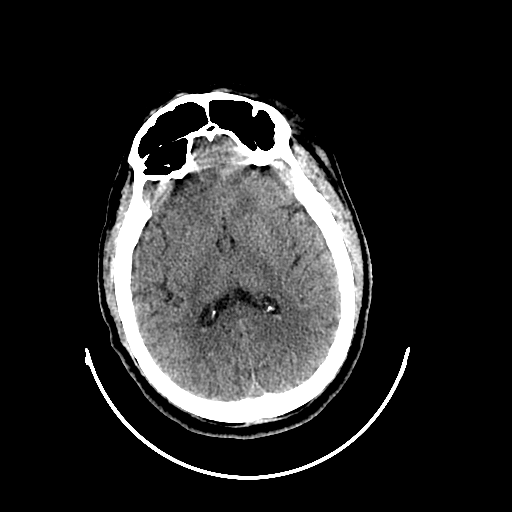}
			\caption{False negative: tiny intraventricular hemorrhage}
		\end{subfigure}
		\hfill
		\begin{subfigure}{0.3\textwidth}
			\includegraphics[width=\textwidth]{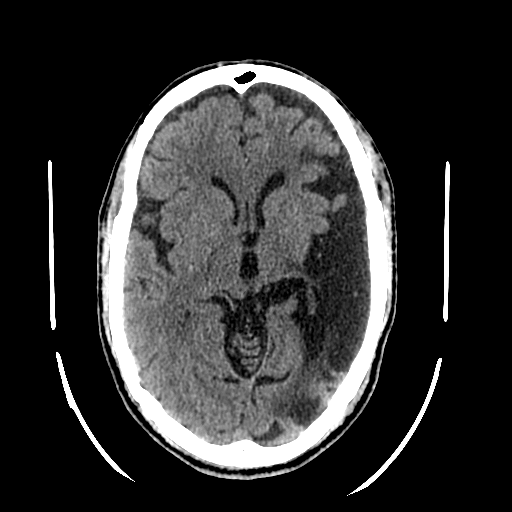}
			\caption{False positive: predicted as subdural hemorrhage}
		\end{subfigure}
		\hfill
		\begin{subfigure}{0.3\textwidth}
			\includegraphics[width=\textwidth]{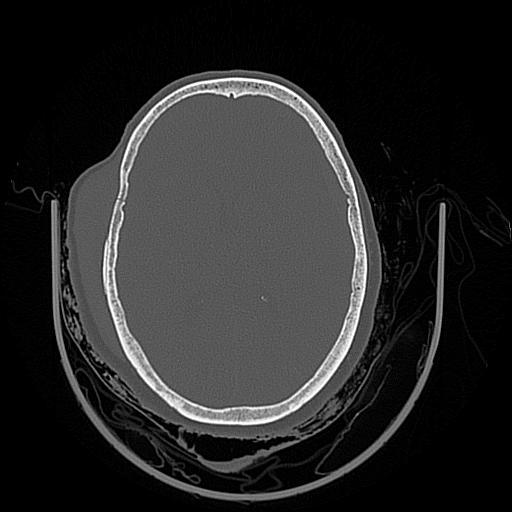}
			\caption{False negative: calvarial fracture (in right parietal bone)}
		\end{subfigure}
		\caption*{Erroneous predictions}
		\vspace{3mm}
	\end{subfigure}
	\caption{Some accurate and erroneous predictions of the algorithms}
	\label{fig:examples}
\end{figure}

We have done an informal qualitative analysis by going through the algorithms' output and the three readers' opinions on the CQ500 dataset.
It was observed that the algorithms produced good results for normal scans without bleed, scans with medium to large sized intraparenchymal and extraaxial hemorrhages, hemorrhages with fractures and in predicting midline shift.
There was room for improvement for small sized intraparenchymal, intraventricular hemorrhages and hemorrhages close to skull base.
In this study, we didn't separate chronic and acute hemorrhages.
This resulted in occasional prediction of scans with infarcts and prominent CSF spaces as intracranial hemorrhages.
Future research can improve the algorithms to mitigate these failure modes.
We show some accurate and erroneous predictions of the algorithms in Figure \ref{fig:examples}.

Our study has several limitations.
Although the selection strategy ensured that there were a significant number of positive scans in the CQ500 dataset for most of our target findings, number of scans with extradural hemorrhage were found only to be 13.
This makes the results of extradural hemorrhage on the CQ500 dataset somewhat statistically insignificant.

For the scans in the CQ500 dataset, concordance between the three radiologists was not very high for all the findings (see Table \ref{table:concordance}).
In particular, fracture had low Cohen's kappas of $0.5771$, $0.3704$ and $0.3637$ between the pair of readers.
This might be because of non-availability of clinical history to the readers.
We observed that the readers were either very sensitive or very specific to a particular finding.
For example, two readers were highly sensitive to calvarial fracture while the third reader was highly specific. 
Sensitivities and specificities of the readers for intracranial hemorrhage, calvarial fracture and midline shift are shown in Table \ref{table:reader_measures}.

\begin{table}
	\begin{tabularx}{\textwidth}{p{10em} XX XX XX}
		\toprule
		\textbf{Finding} &
		\multicolumn{2}{c}{\textbf{Reader 1}} &
		\multicolumn{2}{c}{\textbf{Reader 2}} &
		\multicolumn{2}{c}{\textbf{Reader 3}} \\

		 &
		Sensitivity &
		Specificity &
		Sensitivity &
		Specificity &
		Sensitivity &
		Specificity \\

		\midrule

		Intracranial hemorrhage &
		0.9805 & 0.8986 &
		0.9268 & 0.9790 &
		0.9122 & 0.9790 \\

		Calvarial fracture &
		1.0000 & 0.9519 &
		0.9706 & 0.9628 &
		0.4118 & 0.9978 \\

		Midline shift &
		0.8769 & 0.9883 &
		0.9385 & 0.9038 &
		0.8000 & 0.9883 \\

		\bottomrule
	\end{tabularx}
	\medskip
	\caption{Sensitivities and specificities of the readers versus the majority vote}
	\label{table:reader_measures}
	\vspace{-0.5em}
\end{table}

Another limitation of our study is that we have not excluded follow up scans of a patient from the CQ500 dataset. This is primarily due to the fact that there were very few scans reported with some of our target abnormalities like extradural and intraventricular hemorrhages.
We couldn't present the extent of this limitation because of non-availability of unique identifier of the patients in this dataset. However, note that there was no overlap of patients or scans between the CQ500 dataset and the development dataset.

\begin{figure}
	\centering
	\begin{subfigure}{0.47\textwidth}
		\includegraphics[width=\textwidth]{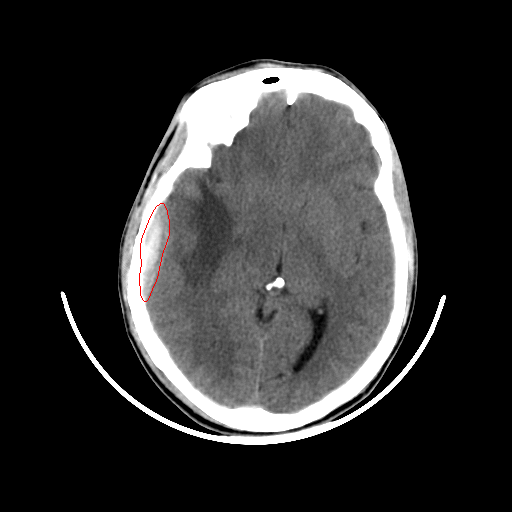}
		\caption{Output produced by the hemorrhage segmentation algorithms.}
	\end{subfigure}
	\hfill
	\begin{subfigure}{0.47\textwidth}
		\includegraphics[width=\textwidth]{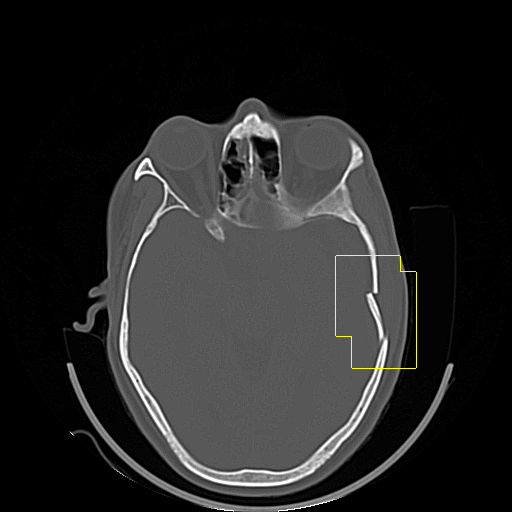}
		\caption{Localization output produced by the calvarial fracture detection algorithm.}
	\end{subfigure}
	\caption{Localizations produced by the algorithms. These can provide a visual display of the results.}
	\label{fig:localization}
\end{figure}

Since we also trained segmentation networks for hemorrhage detection algorithms, we can also output a mask representing the precise location and extent of the hemorrhage (except for the subarachnoid hemorrhage), in addition to detecting its presence.
Similarly, our calvarial fracture detection algorithm can produce a localization output. 
These outputs are represented in Figure \ref{fig:localization}. However in this paper, we have not quantified their accuracy through independent radiologist reads.
In this study, we have limited our algorithm to the calvarial (cranial vault) fractures; the success of the algorithm on detecting these means that our ongoing research can include an extension of the algorithm to all other cranial and facial fractures.

\section{Conclusion}

Our results show that deep learning algorithms can be trained to detect critical findings from head CT scans with good accuracy.
The strong performance of deep learning algorithms suggest they could be a helpful adjunct for identification of acute Head CT finding in a trauma setting, providing a lower performance bound for quality and consistency of radiologic interpretation. 
It could also be feasible to automate the triage process of Head CT scans with these algorithms.
However, further research is necessary to determine if these algorithms enhance the radiologists’ efficiency and ultimately improve patient care and outcome.

\bibliographystyle{unsrtnat}
\bibliography{main}

\end{document}